\documentclass{article}

\usepackage{arxiv}

\usepackage[utf8]{inputenc} % allow utf-8 input
\usepackage[T1]{fontenc}    % use 8-bit T1 fonts
\usepackage{amsmath}
\usepackage{amssymb}
\usepackage{amsfonts}
\usepackage{amstext}
\usepackage{comment}
\usepackage{graphicx}
\usepackage{tipa}
\usepackage[parfill]{parskip}
\usepackage{subcaption}
\DeclareMathOperator*{\argmax}{arg\,max}
\usepackage{hyperref}       % hyperlinks
\usepackage{url}            % simple URL typesetting
\usepackage{booktabs}       % professional-quality tables
\usepackage{amsfonts}       % blackboard math symbols
\usepackage{nicefrac}       % compact symbols for 1/2, etc.
\usepackage{microtype}      % microtypography
\usepackage{amsmath}
\usepackage{lipsum}         % Can be removed after putting your text content
\usepackage{graphicx}
\usepackage{natbib}
\usepackage{doi}
\usepackage{multirow}
\usepackage{algorithm}
\usepackage{comment}
\usepackage{algorithm,algpseudocode}
\usepackage{amsmath}

\usepackage{cleveref}
\algdef{SE}[SUBALG]{Indent}{EndIndent}{}{\algorithmicend\ }%
\algtext*{Indent}
\algtext*{EndIndent}

\title{MAC: A Meta-Learning Approach for Feature Learning and Recombination}

\author{ 
	Sambhavi Tiwari \\
	Department of Information Technology\\
	Indian Institute of Information Technology\\
	Allahabad \\
	\texttt{rsi2018503@iiita.ac.in} \\
    \And
	Manas Gogoi \\
	Department of Information Technology\\
	Indian Institute of Information Technology\\
	Allahabad \\
	\texttt{pcl2017001@iiita.ac.in} \\
    \And
	Shekhar Verma \\
	Department of Information Technology\\
	Indian Institute of Information Technology\\
	Allahabad \\
	\texttt{sverma@iiita.ac.in} \\
    \And
	Krishna Pratap Singh \\
	Department of Information Technology\\
	Indian Institute of Information Technology\\
	Allahabad \\
	\texttt{kpsingh@iiita.ac.in} \\
}

\renewcommand{\shorttitle}{\textit{arXiv} Template}

\hypersetup{
pdftitle={MAC: A Meta-Learning Approach for Feature Learning and Recombination},
pdfsubject={q-bio.NC, q-bio.QM},
pdfauthor={David S.~Hippocampus, Elias D.~Striatum},
pdfkeywords={Meta-learning, few-shot-learning, feature-reuse, optimization-based learning},
}

\begin{document}
\maketitle

\begin{abstract}
	Optimization-based meta-learning aims to learn a meta-initialization that can adapt quickly a new unseen task within a few gradient updates. Model Agnostic Meta-Learning (MAML) is a benchmark meta-learning algorithm comprising two optimization loops. The outer loop leads to the meta initialization and the inner loop is dedicated to learning a new task quickly. ANIL (almost no inner loop) algorithm emphasized that adaptation to new tasks reuses the meta-initialization features instead of rapidly learning  changes in representations. This obviates the need for rapid learning. In this work, we propose that contrary to ANIL, learning new features may be needed during meta-testing. A new unseen task from a non-similar distribution would necessitate rapid learning in addition to the reuse and recombination of existing features. We invoke the width-depth duality of neural networks, wherein we increase the width of the network by adding additional connection units (ACUs). The ACUs enable the learning of new atomic features in the meta-testing task, and the associated increased width facilitates information propagation in the forward pass. The newly learned features combine with existing features in the last layer for meta-learning. Experimental results confirm our observations. The proposed MAC method outperformed the existing ANIL algorithm for non-similar task distribution by $\approx$ 12\% (5-shot task setting).

\end{abstract}

% keywords can be removed

\section{Introduction}
Artificial Intelligence transforms diverse fields with innovation and efficiency such as healthcare \cite{wozniak2023deep}\cite{wozniak2023bilstm}, finance\cite{abe2018deep}, autonomous vehicles\cite{bojarski2016end}, NLP\cite{vaswani2017attention}, object detection\cite{wozniak2022deep}, cybersecurity\cite{ambalavanan2020cyber} and others.
In AI, replicating human adaptability is pivotal. Deep learning performs tasks well but lacks human-like adaptability. A child is able to swiftly adapt to new images, while a deep learning model needs a large amount of training data to learn. The model often overfits and fails to generalize when presented with just one or few new images. This disparity showcases AI's limitations in adapting like humans. Meta-learning presents a solution that enables models to generalize from prior experiences, akin to human adaptability. It exploits the ability of the model to quickly adapt to the new set of data points known as a task \cite{koch2015siamese}\cite{vinyals2016matching}\cite{snell2017prototypical}\cite{finn2017model}\cite{santoro2016meta}\cite{ravi2016optimization}\cite{nichol2018first}. These methods define a family of tasks from a single distribution, some of which are used for training and the rest reserved for evaluation. Meta Agnostic Meta-Learning (MAML)\cite{finn2017model} is the benchmark algorithm for all optimization-based meta-learning algorithms. It works on the principle of two-level, few-shot learning. First, \textit{base learner}, consists of a base module, performs rapid learning from a few-shot task. Second, \textit{Meta learner}, consists of a meta module that optimizes the base learner using unseen meta-test tasks. Raghu et al. \cite{raghu2019rapid} hypothesize that we can obtain the same rapid learning performance of MAML through feature reuse only. The paper contends that MAML learns new tasks by updating the head (the last fully connected layer) with almost the same features (the output of the penultimate layer) from the meta-initialized network \cite{raghu2019rapid}. However, what would happen if the meta-testing task is not from the learned distribution? Will there be new feature learning or just reuse?   \\
To find out what is necessary for meta-learning to happen in the context of \textit{rapid learning, feature reuse} and \textit{ feature learning}, we propose and evaluate MAC (Meta-learning using additional connections) algorithm.\\

Following are the salient contributions of our method which uses extra connections in meta-initialized model to improve how additional features are learned and combined, which enhances adaptability:
\begin{itemize}
    \item {MAC makes use of the width-depth duality in neural network to preserve previously learnt meta-trained features and easily integrate newly learned task-specific features.}
    \item {The method ensures that previously learned atomic and abstract properties are retained by freezing neural network parameters. }
    \item {ACUs enable a focused strategy for new feature extraction. This selective extraction guarantees that gradient updates affect only new connections while preserving the integrity of pre-existing meta-training parameters.}
\end{itemize}

\section{Related Work}

The main motive of meta-learning is to learn across-task prior knowledge to adapt to specific unseen tasks \cite{bengio1995optimization}\cite{hochreiter2001learning}. Meta-learning techniques is divided into three categories:\\

\textit{Memory-Based :} These methods encode fast adaptation into network architecture by generating input conditioned weights or adding an external memory in the network\cite{santoro2016meta}\cite{munkhdalai2017meta}. A number of algorithms\cite{santoro2016meta}\cite{tiwari2022meta} use this method to create a more robust meta-learning model.\\

\textit{Metric-Based :} Learning distance functions or similarity metrics is at the heart of most metric-based approaches. Prototypical Networks \cite{snell2017prototypical} compute class centroids, aiding in fast adaptation to new classes. Matching Networks \cite{vinyals2016matching} utilize a learned similarity metric between support and query instances. Siamese Networks \cite{koch2015siamese} learn embeddings for comparison in a shared space. Relation Networks \cite{sung2018learning} model relationships between samples for improved classification. These methods learn the relationship between support and query data points by defining an embedding space where the same class data points are clustered. In contrast, different class data points are held further apart.\\
    Several advancements in metric-based meta-learning have surfaced. By introducing meta-regularization and self-calibrated inference, Blockmix \cite{tang2020blockmix} improves resilience in few-shot circumstances. Moreover, \cite{peng2019few} has shown that knowledge transfer greatly enhances few-shot image identification. By updating the knowledge of pre-trained models, this technique makes recognition possible with a small amount of labeled data. However, when it comes to problems with little labeled data, integrating external knowledge into the model can facilitate better generalization and adaptation to new classes or tasks with minimal labeled data \cite{li2023knowledge}. Hence, such recent methods dramatically improve the meta-learning models' adaptability and generalization."\\

\textit{Optimization-Based :} For quick adaptation, model-based meta-learning algorithms use optimization to determine the task-specific parameters\cite{ravi2016optimization}\cite{nichol2018reptile}\cite{nichol2018first}. Of all the strategies, MAML \cite{finn2017model} and its first-order approximation is the most widely used. Another algorithm, Meta-SGD\cite{li2017meta}, enhances meta-learning performance by modifying the optimization procedure itself. \\
Recently, some articles \cite{chen2019closer}\cite{raghu2019rapid}\cite{tian2020rethinking} gained popularity by demystifying MAML's success. They proved that learning good features during meta-training and performing rapid learning on new tasks during the meta-adaptation phase is not true. Instead, reusing the learned features during adaptation is the dominant success factor. Among all, ANIL\cite{raghu2019rapid} contends it to be just feature-reuse by providing all necessary experimental results. \\
It is evident from this \cite{raghu2019rapid} research that there is no scope for new feature learning during meta-adaptation phase. Therefore, when the model encounters new features coming from a a different or perturbed task distribution compared to the meta-training distribution, it may struggle to generalize to these unseen tasks, posing difficulties due to the lack of exposure during meta-training.\\

In this paper, we propose a solution to this problem by increasing the width of the base network. Therefore, additional connection units (ACUs) are added to the base model. These units make a provision for learning new features present in the meta-test tasks and recombine them with meta-trained features in a rapid manner. Therefore, the proposed MAC algorithm modifies the ANIL [19] algorithm to learn new atomic features during meta-testing, preserving and using the previously meta-learnt features.
\section{Problem Definition}

The problem focusses on the limitations of existing meta-learning algorithms during the meta-testing phase when it encounters novel tasks with probability distribution that deviate from the original task distribution. Meta-learning algorithms assume consistency between training and test task distributions. However, in real-world scenarios meta-test set's task distribution are often perturbated. In the domain of meta-learning, the challenge arises during the meta-testing phase when novel task distribution and meta training task distribution diverge.  This divergence necessitates a balance between leveraging existing meta-trained features and acquiring new ones. Current meta-learning algorithms struggle to effectively manage this balance, leading to suboptimal learning of changes in the representations when faced with shifted task distributions especially with meagre amount of meta test data.
The fundamental challenge stems from the limitation in existing methods to effectively merge established feature recombination with acquiring new atomic features. This merger is essential, especially for tasks derived from perturbed meta-test distributions.Addressing this shortfall is critical and requires a meta testing phase that can adeptly combine efficient learning of new features and reuse of features already learnt in the meta training phase. A technique that allows the retention and reuse existing with new feature learning  is pivotal for improved adaptation in meta-testing scenarios marked by distributional shifts.

\section{Background Information}
\label{sec:headings}
To understand the concept of feature learning, re-use, and adaptation in meta-learning, we should have some prior knowledge of meta-learning foundation, MAML algorithm, and ANIL algorithm to justify the reason behind why feature re-use is prominent in optimization-based meta-learning approach and how only feature re-use will not help for capturing new atomic features of the perturbed data during the meta-adaptation phase.

\subsection{\textbf{Meta-learning Foundation}}The generic Meta-learning algorithms work on the principle of learning to learn. They works on reusing the learned information known as prior to adapt quickly to the new tasks. 
Meta-learning technique solves the meta-objective (equation 1) to find an optimal meta-parameter $\theta^{*}$ using meta-training dataset $D^{tr}$ with randomly initialized parameters $\theta$.

\begin{equation}
    \theta^{*}=\argmax _{\theta} p(\theta \textpipe  D^{tr})
\end{equation}

\subsection{\textbf{Model Agnostic Meta-learning}}
The \textit{Optimization-based} meta-learning algorithms fine-tunes the model using gradient-based learning rule for a new task($T$) that can make rapid learning. Similarly, MAML model \textit{f} with parameter $\theta$ is trained on multiple tasks to learn the prior and then adapt to the new class task.
Given data $D$ = $\{{D^{tr}},{D^{ts}}$\} drawn from a distribution $p$(${D}$), where ${D^{tr}}$ refers to meta-training dataset and ${D^{ts}}$ refers to meta-testing dataset. In order to perform meta-training, we sample $m$ batch of tasks ${T}_m= \{T_1,T_2,...,T_m$\} from meta-train dataset ${D^{tr}}$. \\
For meta-testing, we sample $n$ batch of unseen tasks \{$ T_1,T_2,...,T_n$ \} from meta-test dataset ${D^{ts}}$. Each meta-training task $T_i=\{{(x_{1},x_{2},...,x_{S}), (x_1,...,x_{Q})} \}$ have $S$ number of labelled data points known as \textit{support set} which are used for inner-loop updation and $Q$ number of labelled data points known as a \textit {query set} used to update the outer loop to a position that produces optimal parameters for quick adaptation. \\

For each inner loop updates, for each task $T_{i}$ we compute \\
\begin{equation}
    \theta_{j}= \theta_{j-1} -\alpha \nabla_{\theta_{j-1}}\mathcal{L}_{T_i}
(f_{\theta_{j-1}})
\end{equation}
Where $j = \{1,2,...,k\}$ is fixed num-steps for every task and $\mathcal{L}_{T_i}
(f_{\theta_{j-1}})$ is the loss on $S$ support set of task $T_i$ after $j-1$ inner gradient step update.
The meta-loss for $M$ batch of task will be:
\begin{equation}
    \mathcal{L}_{meta}(\theta) = \sum_{m=1}^M \mathcal{L}^{Q}_{T_{i}} (f_{\theta_{j}})
\end{equation}
where $\mathcal{L}^{Q}_{T_{i}} (f_{\theta_{j}})$ is the loss computed on querry set of task $T_i$ after $j$ inner loop update. Finally, the outer loop finally updates $\theta$ to $\theta^{*}$:
\begin{equation}
\theta^{*} = \theta - \eta {\nabla_{\theta}} \mathcal{L}_{meta}(\theta)     
\end{equation}

To, perform meta-testing draw test-task $T_i$ from $D^{ts}$ distribution to find loss $\mathcal{L}^{Q}_{T_{i}} (f_{\theta_{j}})$ after inner-loop update using support set $S_{T_i}$ of test task and outer learning rate $\nabla$. Then we compute the accuracy on the querry set $Q_{T_i}$ of test task. 

MAML\cite{finn2017model} learns via two optimization loops:\\
\\1. \textit{Outer loop:} This loop is dedicated to finding the meta-initialization parameters, i.e., from $\theta$ to $\theta^{*}$.\\ Where $\theta$ is the randomly initialized parameters and $\theta^*$ is the meta-initialization parameters obtained after each task inner loop update.
\\2. \textit{Inner loop:} Initialises with outer loop parameters and for each task performs few-gradient updates on $k$ labelled data points (support set) to perform task-adaptation.

\subsection{\textbf{Almost No Inner Loop}}
ANIL \cite {raghu2019rapid} tries to prove that MAML solves the new unseen tasks by feature reuse, not rapid learning. It shows that it achieves the same performance as MAML with feature reuse in optimization-based meta-learning. Apropos of MAML, feature reuse means little or no adaptation for the inner loop during a meta-training and meta-testing phase.
For all feature layers, the CCA metric computed is almost unity; there is virtually no change during adaptation, indicating only feature reuse. In addition, they found that this feature recombination is seen during the early meta-training phase.\\
\\The ANIL model consists of '$L$' layers where $\{1,2,3,4..., L-1\}$ layers are the hidden layers of the networks denoted as \textit{body} of the network and $L^{th}$ layer is called as \textit{head} of the network.
After meta-training, we obtain meta-initialization parameters $\{\theta_1,\theta_2,\theta_3,...,\theta_L\}$ for $L$ layers. For every test task $T_i$, perform meta adaptation using $k$ inner gradient steps using following update rule:
\begin{equation*}
    \theta {_{k}^{(i)}} = (\theta_1,\theta_2,,...,\theta {_{k-1}^{(i)}} - \alpha\nabla_{(\theta_L){_{k-1}^{(i)}} }\mathcal{L}_{S_i} (f_{\theta{_{k-1}^{(i)}}}) ) 
\end{equation*}
Where $S_i$ is the support set of meta-test task $T_i$ and $\alpha$ is the inner-loop learning rate.\\
The update mentioned above clearly re-uses the meta-initialised parameters of all \{$L-1$\} layers except the final layer $L$ and the results\cite{raghu2019rapid} show that this update is similar to MAML. ANIL convincingly proved that MAML does feature re-use rather than rapid learning.

\section{\textbf{M}eta-\text{L}earning using \textbf{A}dditional \textbf{C}onnection \textbf{U}nits} \label{sec5}
The MAC algorithm is a solution to the problem of learning new data features when meta-train and meta-test data have slightly dissimilar distributions. This is achieved by one, increasing the width of the base network and two, by integrating ACUs into the model. These ACUs enable efficient and rapid learning of new features in meta-test tasks and incorporate them with previously learned features. The method uses the technique in ANIL to learn and re-combine new atomic features during meta-testing.

We invoke the breadth-depth duality of neural networks, wherein we increase the width of the network by ACUs. The depth is kept almost unchanged so that features learnt during meta training remain fixed. Freezing the neural network parameters post-meta-learning phase allows us to retain the atomic and abstract features learnt in the meta-training phase. Concurrently, we enable combining all features in the higher layers for meta-learning in the meta-test phase. The method allows feature recombination as the gradient update during the inner loop is restricted only to change the weights of the additional connections while keeping the meta-train parameters unchanged. This holds the prior generated during meta-training and accommodates learning of new features in the meta-testing phase.\\

\textbf{Width and Depth in Neural Networks :} In a neural network, there is a width-depth duality and either depth or width can enable a sufficient representation ability \cite{fan2020quasi} \cite{nguyen2020wide}. As a task becomes complicated, the width and depth must be increased accordingly to promote the expressive power of the network. Increasing the width is essentially equivalent to increasing the depth for boosting the hypothesis space of the network\cite{nguyen2017loss}. It can be hypothesized \cite{fan2020quasi} \cite{nguyen2020wide} \cite{nguyen2017loss} that for finer details, wider layers are important and depth emphasizes global structure. 

The proposed N-way classification setup is as follows:\\
We initialise the base model \textbf{M} to be a neural network ($f_\theta$) with  $\theta$ as initial parameters. The model is divided in two parts i.e., a) All the layers except the last layer of the network is termed as the \textbf{body}, and b) the classifier layer of the network is called the \textbf{head}. Given data $D = \{D^{tr},D^{ts}\}$, draw meta-training batch of m tasks ${T}_i= \{{T_1,T_2,...,T_m}\}$ from $D^{tr}$ and meta-testing batch of n tasks $T'_{i}= \{{T'_1,T'_2,...,T'_n}\}$ from ${D^{ts}}$ . Notably, $p({D^{tr}})$ is the original task distribution and ${p({D^{ts}})}$ is the perturbed task distribution obtained after adding \textit{Gaussian} noise to $P(D)$. Each task ${T}_i$ = \{$S_{T_i}$,$Q_{T_i}$\}  has \textit{support set} $S_{T_i}$ = \{${x_j,y_j}$\}  and \textit{query set} $Q_{T_i}$ =\{${x_k,y_k}$\} .Similarly, for meta-test task $T'_{i}$ =  \{$S_{T'_i}$,$Q_{T'_i}$\} has \textit{support set} $S_{T'_i}$ = \{${x'_j,y'_j}$\}  and \textit{query set} $Q_{T'_i}$ =\{${x'_k,y'_k}$\}.\\

Next, we modify the base model $M$ to MAC model $M'$ to perform meta-testing on meta-test batch of tasks $T'_{i}$. 

\subsection{MAC Architechture}
\begin{figure*}[ht]
\centering
\includegraphics[width=\linewidth]{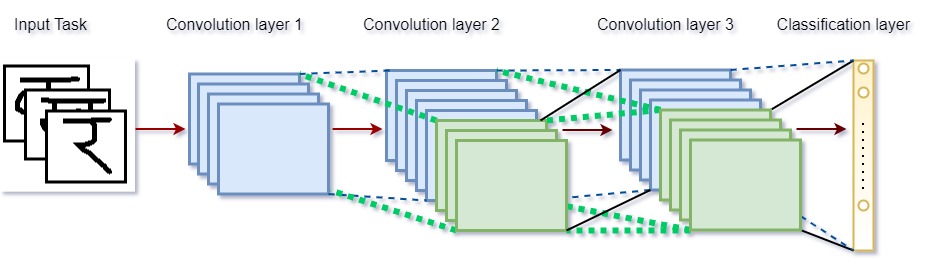}
\caption{This diagram shows the proposed MAC model $M'$. The neural network inputs a batch of task and modifies the weights of the networks in each iterations. Each convolution layer has blue and green filters. The blue filters are the base modules and green filters are the newly added units(ACUs). Blue dashed lines are the old connection of the base model \textbf{M}, Whereas green dotted connections are the new links with the old and new nodes. Solid black lines are zero weight connections.    }
  \label{fig:1}

\end{figure*}

The new model $M'$ (Figure.1) is formed by adding \textbf{additional connection units} in the body of the base model $M$. An illustration of the proposed network model $M'$ for meta-testing is shown in fig.1. For each $'l'$ layer \{${1,2,3, . . . ,L}$\} of the network $f$ takes input as an image of the task and outputs a class label for the corresponding task image. The \textbf{ACUs} are added on $\{L-1\}$ layers of the network. Let $\theta^* = (\theta_{1}^*,\theta_{2}^*,...,\theta_{L}^*)$ be the meta-trained parameters for the each layer of the network.  We initialise layer weights of the MAC model $M'$ with $\theta^*$.\\
For $L-1$ layers: The weight matrix ${W_l}$ for the body of the network  \{${l=1,2,...,L-1}$\} is initialised as:\\

\begin{equation}
W_{l} = 
\begin{pmatrix} 
\theta^{*}_{l} & 0 \\
W_{l1} & W_{l2} 
\end{pmatrix},\\
b^l = 
\begin{pmatrix}
b^{*}_{l}\\
b_{l}
\end{pmatrix}
\end{equation}
\\
where,${W_{l1}}$ are weights w.r.t the connections between ACU and base layer. ${W_{l2}}$ are new weights w.r.t the connections between ACUs. Both ${W_{l1}}$ and ${W_{l2}}$ are randomly initialized weights. The zero weights are initialised for the connection between new to old nodes of the network.\\  
Similarly, $b_{l}$ is randomly initialised bias for ACUs and 
$b^{*}_{l}$ is the meta-trained bias obtained during meta-training.

For $L^{th}$ layer: The weights and bias matrices for classifier layer of model $M'$ will be initialised as:\\

\begin{equation}
W_{L} =
\begin{pmatrix}
\theta^{*}_{L} & 0 
\end{pmatrix}   ,
b_L = 
\begin{pmatrix}
b^{*}_{L}
\end{pmatrix}
\end{equation}

where, $\theta^{*}_{L}$ is the meta-trained parameters of the classifier layer and zero weights are the connection weights from newly added ACUs to classifier layer nodes.  

\subsection{\textbf{MAC Training}}
 Meta-training is performed on a batch of $m$ tasks drawn from the training data distribution $p(D^{tr})$. $\theta$ be the initial parameters for [L-1] layers, and $\mu$ be the initial classifier layer parameter of the base model $M$. Initially, both the parameters are randomly initialized.   The process of Meta-training for MAC and ANIL\cite{finn2017model} algorithm is same. For each batch of task, we perform two updates:\\
 
1) In inner loop gradient updates, we only update the classifier layer weights $\mu$ of the proposed model $M$.\\
2) During the outer loop update, we calculate and update the gradients of the model $M$ after each inner loop updation.\\

Finally, we get our meta initialized parameters $\theta^*$ for [L-1] layers and $\mu^*$ for the classifier layer. Note that $\mu$ is updated in both the inner and outer loop, but $\theta$ is updated only during the outer loop.

\begin{figure*}[ht]
\centering
\includegraphics[width=\linewidth]{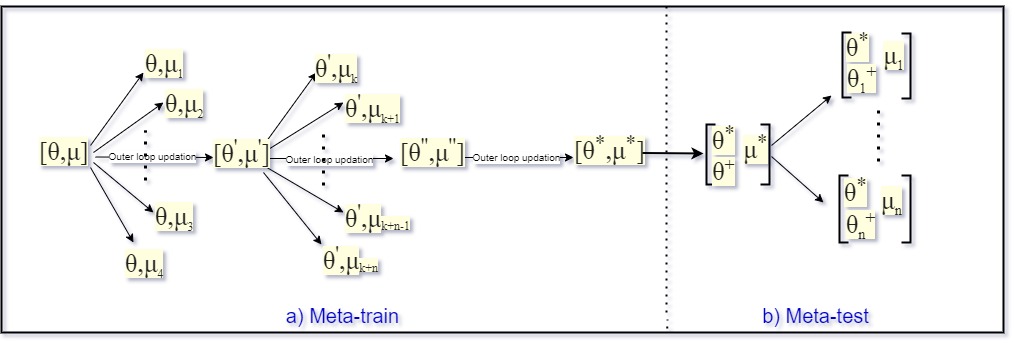}
\caption{Part (a) illustrates the meta-training phase of our proposed MAC algorithm where it finds meta-initialization parameters $\theta^*$ and $\mu^*$ after the final outer loop updation for the model $M'$ so that new tasks are learnt rapidly with additional feature learning and recombination, part (b) Represents model $M'$ with additional links to the [L-1] layers shown as ${\theta^+}$ and the classifier layer weights as $\mu^*$ obtained during meta-training to perform task adaptation. }
\centering
\end{figure*}

\subsection{\textbf{MAC Testing}}

Fig. 2(b). illustrates the meta-testing phase on model $M'$ for the data whose distribution is not similar to the training data distribution. To achieve this, we update our model $M$ by adding extra links in the [L-1] layers. These links are ACUs and are represented as $\theta^+$ in the diagram. The model is initialized with meta-trained parameters $ \theta^*$  and $\mu^*$. Whereas $\theta^+$ are the randomly initialized parameters of newly added ACUs. These additional links in the network are designed to capture new atomic features of the perturbed tasks, which then recombine with the meta-trained parameters and perform few-shot-learning. During meta-testing, each task updates the ACU parameter($\theta^+$) and the classifier layer($\mu^*$) parameters to perform meta-adaptation on the new unseen tasks. \\

For a clearer view, we will see how the parameters are updated during forward and backward pass.\\
a) \textbf{{Forward pass}}: Refer to Fig. 1 and 2, the model $M'$ has two coloured filters in the last two hidden layers(the number of hidden layers may vary). Blue colored filters are meta-trained parameters($\theta^*,\mu$) of model \textbf{M} and green filters are the parameters of ACU($\theta ^ +$). During forward pass, the image of task \textbf{T'} is passed through each layer of model $M'$. An extended feature map is obtained by the newly added ACUs and propagated to the next layer during the forward pass.\\
b) \textbf{{Backward pass}}: In backward pass,gradients are backpropagated through a subset of connections only (Fig.\ref{fig:1} : green dots). On the other hand, certain gradients are first purposefully made \textit{zero} (solid black lines). Later on, these zeroed-out weights will be modified in the meta-adaptation phase. This will essentially make the model learn the new atomic features and just update the classifier layer and ACU parameters.\\

During Meta-testing, the ACUs learn new features and recombine them with learned features. A thorough description of the MAC testing is explained in Algorithm.\ref{algo1}.

\begin{algorithm}
    \caption{Algorithm for meta-testing}
    \renewcommand{\algorithmicrequire}{\textbf{Input}: ${p(\mathcal{T'})}$ distribution over meta-test tasks,}
    {\textbf{Initialize:}}'$\alpha$' learning rate, '\textit{n}' step size, size of \textbf{ACU}
    \begin{algorithmic}[1]
        \State Initialize $\theta^*$ and $\mu^*$ from meta-traininig.
        \State Sample a batch of tasks ${T'}_{i}$ from ${D^{ts}}$
 
        \For {all \textbf{i} } 
            \State Sample $K$ datapoints $\{{{x'}_{j}, {y'}_{j}}\}$ from $\mathcal{T'}_{i}$(N-way K-shot).
            \State Initialize additional connection units as ${\theta^+} \sim (\mathcal{N}(0,1))$ :
            
            and Let, $\Theta \rightarrow [{\theta^*}, {\theta^+},{ \mu^*}],$ 
             
            \State Evaluate $\nabla _{[\theta^+,\mu^*]} {\mathcal{L}_{\mathcal{T}_i}}$ using stochastic gradient descent for n-steps.
            \State Update the network parameters:
            
            $\Theta' = \Theta - \alpha {\nabla _{[{\Theta^+},{\mu^*}]}} \mathcal{L}_{\mathcal{T}_{i}} ( f_\Theta )$
            \State Compute accuracy:
            
             ${a_i} \leftarrow accuracy ({{T'}_{i}}, \Theta')$
             
            Total accuracy = ${\frac{1}{\mathcal{\textpipe T \textpipe}}} { \sum_{i=1}^{i} a_i}$
        \EndFor
        
    \textbf{return} Total accuracy
    \end{algorithmic}
    \label{algo1}
\end{algorithm}

\section{Implementation Details and Results}\label{sec6}

\subsection{\textbf{System Configuration}}
We used a high-performance two GPU server for our experimental operations, combining the power of two 16GB memory-equipped Tesla V100 GPUs. In addition, the server included 32GB of RAM and 10,240 CUDA cores with CUDA version V9.2.148. Utilizing a 64-bit system with an x64-based CPU, this setup guaranteed quick and stable experiment execution, allowing dependable and effective performance all through our research.

\subsection{\textbf{Dataset}}
We have used two standard benchmark datasets often used for a few shot learning paradigms. The proposed MAC meta-learning technique has demonstrated success on image sizes of 28x28 (omniglot) and 84x84 (Miniimagenet) using available GPU resources.

\subsubsection{Omniglot}
\begin{figure}[ht]

\includegraphics[width=9cm]{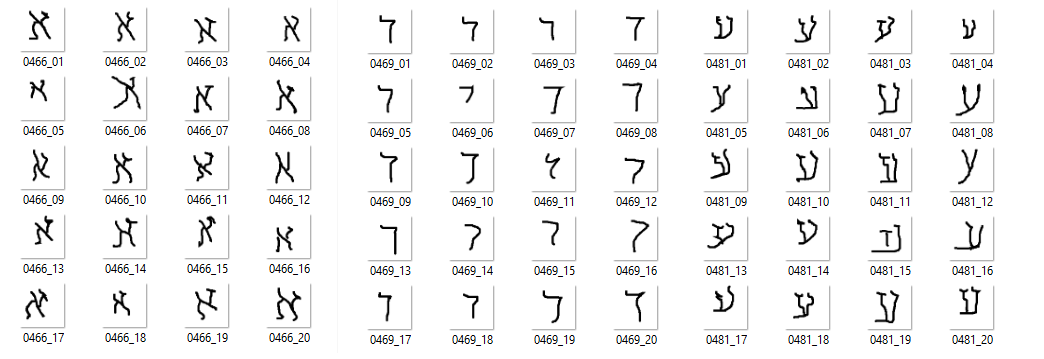}
\centering
\caption{Three sets of different character depicting three classes from Omniglot dataset. ${char1}$, $char4$ and $char16$ in Hebrew. Each character is written by $20$ different person.}
\centering
\end{figure}

To evaluate our proposed model, We perform our experiments on the Omniglot dataset \cite{lake2011one} for classification. This dataset contains 50 different alphabets divided into 1623 handwritten characters known as classes. Each class has 20 black and white images of size 28X28 drawn by 20 different persons. The images are labelled with the name of the corresponding character and a suffix. For, e.g. the Alphabet of the Sanskrit language in the dataset has 42 characters, and there are 20 images of each character with labels. In our classification task, each character is considered a separate class irrespective of language. These classes are split into training and test sets: 1200 classes for training and 423 class for testing. All the character images are first augmented by performing rotations to create more data samples and reduce overfitting. Figure 1 illustrates some Hebrew language characters.
\subsubsection{MiniImagenet}
A subset of the ImageNet collection, with 600 images per category across 100 distinct categories, is called Mini-ImageNet. Every image has a size of 84 x 84. Ravi and Larochelle proposed the MiniImagenet dataset in the year 2016\cite{vinyals2016matching}.  The dataset consists of 64 training classes, 24 test classes and 12 validation classes. We consider four task settings, i.e., 5-way 1-shot, 5-way 5-shot, 10-way 5-shot and 10-way 1-shot on this dataset. Therefore, we have meta-train and meta-test tasks to classify among 20 randomly chosen different classes, given only a few labelled samples, i.e., 5 and 1 instance of each class. The experiments are done for 5-way 1-shot and 5-way 5-shot task settings.

\subsection{\textbf{Implementation Details}}

\subsubsection{Data from non-identical distribution} The main idea is to perform meta-adaptation during meta-testing on the different data distributions to show that the new atomic features are learned using the newly added ACUs. Whereas, in traditional few shot learinng\cite{finn2017model}, task distribution for meta-training and meta-testing are the same. The meta-test tasks are never seen beforehand, and meta-learning assumes it to be from the same distribution as the meta-training task distribution, which is more likely to perform task-adaptation on learned parameters. Past researches performed cross-domain few-shot learning \cite{oh2020boil} \cite{miranda2021does} and used different datasets to perform meta-learning. Unlike others, we created meta-testing data using the actual datasets(omniglot, miniimagenet) and converted it to different distributions using gaussian blur. In terms of method, our work is more similar to ANIL \cite{raghu2019rapid}, and this is because we clubbed our idea of \textit{additional connection units} with existing \textit{almost no inner loop} known as MAC to perform meta-learning for heterogeneous data. \\

\subsubsection{Adding Noise to the dataset}
To perturb the data distribution we chose The \textbf{GaussianBlur} function in PyTorch's \textbf{torchvision.transforms.v2} module. It is a part of the image transformation toolkit. It performs a Gaussian blurring operation on input images by adding noise to the data. In the experiment we used GaussianBlur${(kernel\_size=(5, 9), sigma=(0.1, 5))}$ with two parameter values. The Gaussian kernel size is indicated here by the ${kernel\_size}$ parameter, which ranges from 5 to 9 and the Gaussian kernel's standard deviation is determined by sigma. The higher sigma value results in a blurrier image. The objective of this work is to propose and evaluate a method that is able to adapt to non iid setting. Gaussian noise is
added to the data so that the perturbation changes the underlying data distribution so that it becomes non iid in nature.

\subsubsection{Few-Shot classification}
We used PyTorch and Torchmeta \cite{deleu2019torchmeta} MAML, FOMAML and ANIL implementation.
For a few-shot classification, we used the same model architecture as used in the original MAML implementation \cite{finn2017model}.
To perform N-way k-shot classification, We trained our model using an omniglot dataset for 60000 iterations with a batch size of 32 tasks, three gradient steps, a learning rate of 0.4 and the model is evaluated on a batch size of 32 with step size 0.4. For the Miniimagenet dataset, the model was trained for 60000 iterations, using five gradient steps, batch size of 4 and learning rate of 0.01 and is evaluated using two different learning rates: 0.1 for hidden layers and 0.02 for the classifier layer with ten gradient steps and batch size of 4.\\
\\We performed a comparative evaluation of our MAC technique with three benchmark meta-learning algorithms— MAML, FOMAML, and ANIL specifically in the context of few-shot classification. We evaluated the classification accuracy on 100 batches of unseen tasks using 10 inner stochastic gradient descent steps. We performed several experiments to commit the final observation. We evaluated our model 3 times with different random seeds and calculated the mean of all the results to show the absolute average accuracy. Further, we bound the number of ACUs (Filters for CNN model) added should not be more than 50 units. Since we aim to capture only the atomic features. Therefore, the proposed model enhances the few-shot learning ability that can reuse and recombine the features to adapt to a new task efficiently without complicating the network architecture.
\subsection{Results and Discussion}
The MAC model requires well-defined experimental findings in order to determine the effectiveness of few-shot classification:\\

\textit{Why we need ACUs?}, \textit{How many ACUs are required to learn important atomic features of the new i.i.d. dataset?}, \textit{Where should we add ACUs in our base model?}, \textit{ What happens if we increase the depth of our proposed MAC model?}.
The experimental results and discussion are provided further:

\subsubsection{Need for ACUs}
The main idea behind adding the ACUs is to increase the width of the network. In the meta-learning scenario, it is imperative to learn new atomic features when the meta-testing tasks are drawn from a non-identical distribution. This would require unfreezing parameters of the entire network to enable learning new atomic features in the lower layers, followed by their combination in the higher layers. However, this would lead to unlearning meta-train parameters, effectively stopping the making of the meta-learning process, especially when a non-negligible number of new atomic features needed to be learnt. We can increase the width of the network by adding ACUs to learn new features and reuse the existing features by retaining the learnt parameters to meet the objectives of learning, reuse and recombination. Fig.4.(a) shows the equal learning accuracies for both MAC and ANIL algorithms. This means ACUs in the MAC method do not capture any new features as new atomic features are absent in the same task distribution settings. Whereas fig.4.(b) shows high accuracy for the MAC method, proving that ACUs in the MAC method are responsible for capturing and recombining new extra features during the task adaptation phase.

\begin{figure*}[ht]
\centering
\begin{subfigure}{.45\textwidth}\centering
  \centering
  \includegraphics[width=6.5cm]{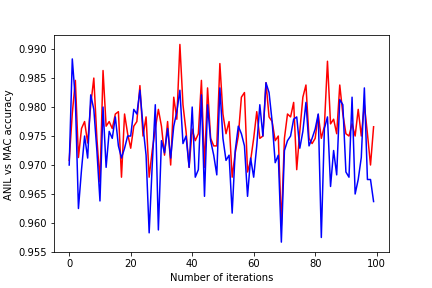}
  \caption{ANIL vs MAC meta-test accuracy on same data distribution(without perturbation).Blue line shows the MAC accuracy and red line shows the ANIL accuracy.}
  \label{fig:sub1}
\end{subfigure}%
\begin{subfigure}{.45\textwidth}
  \centering
  \includegraphics[width=6.5cm]{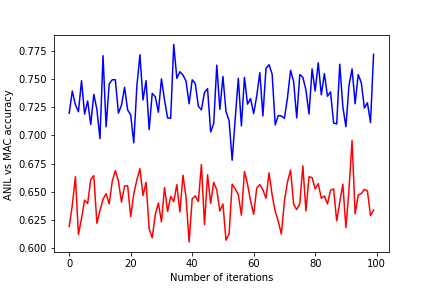}
  \caption{ANIL vs MAC meta-test accuracy on perturbed distribution. Blue line shows the MAC accuracy and red line shows the ANIL accuracy.}
  \label{fig:sub2}
\end{subfigure}
\caption{5-way 5-shot test accuracy graph on omniglot dataset.}
\label{fig:test}
\end{figure*}

\subsubsection{Number of ACUs}
How many node connections (\textit{filters in our case}) should be added to each layer? The size of the ACUs is decided by iterative testing performed for different combinations of nodes for the entire network. We limit our maximum number of nodes to be 50 only. Since we need to extract and recombine the new atomic features of the unseen tasks, we require small connection units. 
However, The size of ACUs is not fixed for all the target task settings, i.e., the size differs for every few-shot task setup. For instance, the optimal number of ACUs added in the model for a 5-way 5-shot task setting is [50, 30, 20, 5], but for a 10-way 5-shot task, it is [50,35,20,10]. The number of ACUs introduced in each layer reflects the need for first layer feature extraction using bigger units and subsequent feature recombination in the later levels utilizing smaller units.\\

After examining every combination of nodes, we found one combination to be optimal for all tasks, i.e.,
\textit{For omniglot}:  $MAC_{opt}$ = [45,35,20,10]. Table I.a shows the experimental results for fixed number of ACUs ($MAC_{opt}$) vs ($MAC_{best}$).
\textit{For miniimagenet}: $MAC_{opt}$ = [50,40,25,20]. Table I.b shows the experimental results for fixed number of ACUs ($MAC_{opt}$) vs ($MAC_{best}$).

\begin{table}[ht]
\centering

\begin{tabular}{|l|c|c|c|c|}
\hline
Method & 5-way 1-shot & 5-way 5-shot  & 10-way 1-shot & 10-way 5-shot \\
\hline
MAML&
42.03&
62.26&
31.91&
48.44\\
\hline
FOMAML&
41.81&
61.70&
31.03&
48.11\\
\hline
ANIL &41.65&
64.43&
30.87&
49.88\\
\hline

$MAC_{best}$&44.12& 74.52
 & 33.42 &57.18\\

\hline
$MAC_{opt}$&43.81& 74.45
 & 33.10 &56.93\\
\hline
\end{tabular}
\subcaption*{Table I.a:  This table shows few shot accuracy ($MAC_{opt}$) vs ($MAC_{best}$). The most optimal number of ACUs are $MAC_{opt}$ = [50,40,25,20] for Omniglot Dataset with Gaussian\_Blur.}
\bigskip

\begin{tabular}{|l||c|c|}
\hline
Method & 5-way 1-shot & 5-way 5-shot \\
\hline

MAML&25.21&
33.66
\\
\hline
FOMAML&24.39&
34.84
\\
\hline
ANIL &25.87 &34.82 \\
\hline
$MAC_{best}$&27.02& 36.12\\
 \hline
$MAC_{opt}$ &26.43&
35.88\\
\hline
\end{tabular}
\subcaption*{Table I.b: This table shows few shot accuracy ($MAC_{opt}$) vs ($MAC_{best}$). The most optimal number of ACUs are $MAC_{opt}$ = [50,40,25,20] for MiniImagenet Dataset with Gaussian\_Blur.}
\end{table}

\subsubsection{Position of ACUs } Our base model $M$ is similar to ANIL\cite{raghu2019rapid} model. The model has 4 modules; each module consists of a 3 x 3 convolution layer, 64 filters with stride 2, followed by a batch normalization layer and reLU at last. We can add a maximum of 4 Additional Connection Units, i.e., one ACU per module, in our base model $M$. To decide where should we add the ACUs, we performed experiments by adding units to each convolution layer and examined the results. Let model $M$ having $L$ layers, and then we add units to at most $L-1$ hidden layers as follows:

a) \textbf{At the beginning of the model:} For our proposed method MAC, we add connections in the initial layers $\{ACU_1, ACU_2,0,0\}$ of the model $M$. We examined the number of optimal ACUs to be added in the initial layers of the model (refer to Table II.a \& II.b ). 

\begin{table}[!ht]
\centering

\begin{tabular}{|l||c|c|c|c|}
\hline
Methods& 5-w 1-s & 5-w 5-s & 10-w 1-s & 10-w 5-s \\
\hline
MAML&
42.03&
62.26&
31.91&
48.44\\
\hline
FOMAML&
41.81&
61.70&
31.03&
48.11\\
\hline
ANIL&
41.65&
64.43&
30.87&
49.88\\
\hline
MAC&43.41& 73.87
 & 32.45 &56.35\\

Units&[25,45,0,0]&[25,50,0,0]&[25,45,0,0]&[25,50,0,0]\\
\hline
\end{tabular}

\subcaption*{Table II.a: Few-Shot Accuracy on Omniglot Dataset with Gaussian\_Blur: Effect of Additional Connections in Initial Layers of the Proposed MAC Model}\label{Tab:1.a}

\bigskip

\begin{tabular}{|l||c|c|}
\hline
Methods& 5-w 1-s & 5-w 5-s \\
\hline
MAML&25.21&
33.66
\\
\hline
FOMAML&24.39&
34.84
\\
\hline
ANIL&26.43&
35.88
\\
\hline
MAC&26.41&36.00 \\

Units&[20,40,0,0]&[45,5,0,0]\\
\hline
\end{tabular}
\subcaption*{Table II.b: Few-Shot Accuracy on MiniImagenet Dataset with Gaussian\_Blur: Effect of Additional Connections in Initial Layers of the Proposed MAC Model}
\end{table}
b) \textbf{At the end of the model $M$:} We added a few ACUs on the last few layers except the classifier layer  $\{0,0, ACU_1, ACU_2\}$ of the model $M$. We further calculated the optimal number of connections to be added in a few last layers of the model ( refer to tables III.a \& III.b).

\begin{table}[!ht]
\centering
\begin{tabular}{|l||c|c|c|c|}
\hline
Methods& 5-w 1-s & 5-w 5-s & 10-w 1-s & 10-w 5-s \\
\hline
MAML&
42.03&
62.26&
31.91&
48.44\\
\hline
FOMAML&
41.81&
61.70&
31.03&
48.11\\
\hline
ANIL&
41.65&
64.43&
30.87&
49.88\\
\hline
MAC&43.62& 74.10
 &31.52 &56.60\\

Units&[0,0,50,20]&[0,0,45,5]&[0,0,50,20]&[0,0,50,5]\\
\hline
\end{tabular}
\subcaption*{Table III.a: Few-Shot Accuracy on Omniglot Dataset with Gaussian\_Blur: Effect of Additional Connections in End Layers of the Proposed MAC Model}\label{Tab:2.a}
\bigskip
\begin{tabular}{|l||c|c|}
\hline
Methods& 5-w 1-s & 5-w 5-s \\
\hline
MAML&25.21&
33.66
\\
\hline
FOMAML&24.39&
34.84
\\
\hline
ANIL&26.43&
35.88\\
\hline
MAC&26.40&35.50 \\

Units&[0,0,40,15]&[0,0,45,5]\\
\hline
\end{tabular}
\subcaption*{Table III.b: Few-Shot Accuracy on MiniImagenet Dataset with Gaussian\_Blur: Effect of Additional Connections in End Layers of the Proposed MAC Model}\label{Tab:2.b}
\end{table}

c) \textbf{Throughout the model $M$:} We added 4 ACUs in each convolution layer except the input and classifier layer  $\{ACU_1, ACU_2, ACU_3, ACU_4\}$ of the model $M$ ( refer to table IV.a \& IV.b).
\begin{table}[!ht]
\centering
\begin{tabular}{|l||c|c|c|c|}
\hline
Methods& 5-w 1-s & 5-w 5-s & 10-w 1-s & 10-w 5-s \\
\hline
MAML&
42.03&
62.26&
31.91&
48.44\\
\hline
FOMAML&
41.81&
61.70&
31.03&
48.11\\
\hline
ANIL&
41.65&
64.43&
30.87&
49.88\\
\hline
MAC&44.12& 74.52
 & 33..42 &57.18\\

Units&[50,40,20,10]&[50,30,20,5]&[50,25,20,10]&[50,35,20,10]\\
\hline
\end{tabular}
\subcaption*{Table IV.a: Few-Shot Accuracy on Omniglot Dataset with Gaussian\_Blur: Effect of Additional Connections in each Layers of the Proposed MAC Model}\label{Tab:3.a}
\bigskip
\begin{tabular}{|l||c|c|}
\hline
Methods& 5-w 1-s & 5-w 5-s \\
\hline
MAML&25.21&
33.66
\\
\hline
FOMAML&24.39&
34.84
\\
\hline
ANIL&26.43
&35.88
\\
\hline
MAC&27.17&38.51 \\

Units&[50,35,15,5]&[50,35,40,10]\\
\hline
\end{tabular}
\subcaption*{Table IV.b: Few-Shot Accuracy on MiniImagenet Dataset with Gaussian\_Blur: Effect of Additional Connections in each Layers of the Proposed MAC Model}\label{Tab:3.b}
\end{table}

\subsubsection{Adding layers to the base model}
In this section, we will find out an answer to a question, i.e., \textit{Will the performance of the proposed method increases if we add some layers to the proposed MAC model?} \\
A promising work\cite{arnold2021maml} done in the past unveils some interesting unknown properties of the MAML algorithm. Their study found that the MAML\cite{finn2017model} is well suited to the depth of the model architecture. Inspired by this, we evaluated our proposed method for the deeper model. To make our model "deep", we add a few Convolution layers to the rearmost of the proposed MAC model $M'$. Further, to evaluate the meta-adaptation performance of this approach, we used two different models. First, the MAC model $M'$ and second, the deep model ${M'}_{deep}$. 
The deep model adds two Convolution layers in the MAC model $M^{'}$. We performed few-shot classification experiments on both the models ($M'$ and ${M'}_{deep}$). Model ${M'}_{deep}$ consists of 6 modules: 4 modules with 3X3 convolutions and 64 filters with stride(2), followed by Batchnorm and a ReLU activation function and 2 additional modules of a Convolution layer, each followed by a Batchnorm and a ReLU activation functions. The model $M_{deep}$ is trained and tested for two datasets (Omniglot and Miniimagenet) on similar hyperparameters (Section 6.3.3).

\begin{table}[ht]
\centering

\begin{tabular}{|l||c|c|}
\hline
Method & 5-way 1-shot & 5-way 5-shot \\
\hline
MAML&
42.03&
62.26\\
\hline
FOMAML&
41.81&
61.70\\
\hline
ANIL&
41.65&
64.43\\
\hline
MAC&44.12& 74.52\\
\hline
$MAC_{deep}$ &43.59&
75.77\\
\hline
\end{tabular}
\subcaption*{Table V.a: Effect of adding depth to the proposed MAC model $M^{'}$. Few-Shot Accuracy on
Omniglot Dataset with Gaussian\_Blur}
\bigskip

\begin{tabular}{|l||c|c|}
\hline
Methods& 5-w 1-s & 5-w 5-s \\
\hline
MAML&25.21&
33.66
\\
\hline
FOMAML&24.39&
34.84
\\
\hline
ANIL&26.43&
35.88\\
\hline
MAC&27.02&36.12 \\
\hline
$MAC_{deep}$ &23.59&
37.31\\
\hline
\end{tabular}
\subcaption*{Table V.b: Effect of adding depth to the proposed MAC model $M^{'}$. Few-Shot Accuracy on
MiniImagenet Dataset with Gaussian\_Blur}
\end{table}

\subsubsection{Effect of shallow vs deep meta-learning model} 
 We observed increased accuracy that is directly due to enhanced meta-learning when depth is increased. The results(table IV.a and IV.b) shows the change in performance metric with the shallow MAC model($M'$) and the deep model $MAC_{deep}$.

Therefore, it is evident from the results that depth facilitates few-shot learning when added to a base model. After repetitive testing, we made an observation that the results were best for a 5-way 5-shot task setting for both omniglot and miniimagenet datasets.

\subsubsection{Model Complexity}
The computational complexity of our proposed model is determined by analyzing the number of operations executed during both forward and backpropagation. In this sub section we will examine these two phases thoroughly:\\

\textbf{Forward Propagation :} The time complexity analysis of a CNN during a forward pass involves the complex calculations required to process input data through its individual layers.
The computational complexities of a CNN consisting of four modules \cite{vinyals2016matching} need to calculated. Each module is designed with certain operations, including a $3\times3$ convolutional layer with $64$ filters, batch normalization, and ReLU activation function.
This analysis specifically focuses on an input image size of dimension $D$.
In general, Let the size of filter be $(N \times N)$, size of image be $(D \times D)$, total number of filters is $F$, and $C$ be the number of channels of the input image. For single module of the base model $M$ the time complexity of convolution operation is calculated as:
$O(N \times N \times C \times F)$, with a $(D \times D)$ input image having per pixel operation cost as $O(D \times D)$. Therefore, Total convolution cost for base model $M$ will be $O(Convolution\_M)= O(N \times N \times C \times F) \times O(D \times D)$ pixels.\\
Since, Batchnorm and ReLu process insignificant cost during forward propagation we will neglect them. 

In MAC algorithm, we modified the model during meta-testing phase by adding extra filters $z$ known as ACUs in each module. These number vary from layer to layer. So, the convolution cost (pixels processed) for one layer will be ${O(Convolution\_M’)= O(N \times N \times C \times (F+z)) \times O(D \times D)}$. Finally, for the base model $M$ having 4 modules we get, $4 \times O(Convolution\_M)$ and for $MAC_{deep}$ model having 6 modules we get, $6 \times O(Convolution\_M’)$.

\textbf{Back Propagation :}  The main computational challenge in backpropagation is calculating the gradients with regard to the network parameters, namely the weights and biases of the convolutional filters. Let weights in the model is denoted as $W$ and bias is denoted as $b$.\\
Base model $M$ : Only the classifier layer $Cl$ is updated. Total gradient update : $gradient(W\_Cl) + gradient (b\_Cl)$\\
MAC model $M^{’}$ : Update only the ACUs ( $z$ filters) and the classifier layer $Cl$. Total gradient update : $gradient(W\_z + W\_Cl) + gradient (b\_z + b\_Cl)$.\\    
$MAC_{deep}$ :  Update only the ACUs ( $z$ filters) ,2 extra layers($l5$,$l6$), and the classifier layer $Cl$. Total gradient update : $gradient(W\_z + W\_l5 + W\_l6 + W\_Cl) + gradient (b\_z + b\_l5 + b\_l6 + b\_Cl)$.

\textbf{Increasing Image Size :} Omniglot often consists of images with smaller dimensions, such as 28x28 pixels, whereas MiniImagenet consists of bigger images, typically measuring 84x84 pixels. An increase in the size of the input immediately affects the amount of pixels that the model processes, resulting in a significant increase in computational load. The input channel will be 3 in the case of Miniimagenet dataset and using equation 7 the total number of parametrers will be three times greater as compared to omniglot dataset.

\section{Conclusion}\label{sec7}
In this paper, we increased the width of the network by adding computational units to make a provision for learning new features present in the meta-test tasks while keeping the parameters of the base network unchanged. This allowed meta-learning when new feature learning during meta-testing is required. The method enabled few shot classifications on perturbed tasks with higher accuracy than methods that preclude new feature learning. Results show that both feature recombination and feature learning are necessary for meta-testing in tasks that are independent but non-identical from the meta-training task distribution. We also discovered that adding new connections should be done in a restricted manner to discourage increased complexity and computational overhead in the model. Also, a gradual decrease in the number of connections moving from the initial to the final layer proved beneficial to the model's performance. The follow-up analysis on the effect of depth showed that it indeed increases the abstraction level around 10 to 12 percent.

\bibliographystyle{unsrtnat}
\bibliography{references}  %%% Uncomment this line and comment out the ``thebibliography'' section below to use the external .bib file (using bibtex) .

\begin{thebibliography}{35}
\providecommand{\natexlab}[1]{#1}
\providecommand{\url}[1]{\texttt{#1}}
\expandafter\ifx\csname urlstyle\endcsname\relax
  \providecommand{\doi}[1]{doi: #1}\else
  \providecommand{\doi}{doi: \begingroup \urlstyle{rm}\Url}\fi

\bibitem[Wo{\'z}niak et~al.(2023{\natexlab{a}})Wo{\'z}niak, Si{\l}ka, and Wieczorek]{wozniak2023deep}
Marcin Wo{\'z}niak, Jakub Si{\l}ka, and Micha{\l} Wieczorek.
\newblock Deep neural network correlation learning mechanism for ct brain tumor detection.
\newblock \emph{Neural Computing and Applications}, 35\penalty0 (20):\penalty0 14611--14626, 2023{\natexlab{a}}.

\bibitem[Wo{\'z}niak et~al.(2023{\natexlab{b}})Wo{\'z}niak, Wieczorek, and Si{\l}ka]{wozniak2023bilstm}
Marcin Wo{\'z}niak, Micha{\l} Wieczorek, and Jakub Si{\l}ka.
\newblock Bilstm deep neural network model for imbalanced medical data of iot systems.
\newblock \emph{Future Generation Computer Systems}, 141:\penalty0 489--499, 2023{\natexlab{b}}.

\bibitem[Abe and Nakayama(2018)]{abe2018deep}
Masaya Abe and Hideki Nakayama.
\newblock Deep learning for forecasting stock returns in the cross-section.
\newblock In \emph{Advances in Knowledge Discovery and Data Mining: 22nd Pacific-Asia Conference, PAKDD 2018, Melbourne, VIC, Australia, June 3-6, 2018, Proceedings, Part I 22}, pages 273--284. Springer, 2018.

\bibitem[Bojarski et~al.(2016)Bojarski, Del~Testa, Dworakowski, Firner, Flepp, Goyal, Jackel, Monfort, Muller, Zhang, et~al.]{bojarski2016end}
Mariusz Bojarski, Davide Del~Testa, Daniel Dworakowski, Bernhard Firner, Beat Flepp, Prasoon Goyal, Lawrence~D Jackel, Mathew Monfort, Urs Muller, Jiakai Zhang, et~al.
\newblock End to end learning for self-driving cars.
\newblock \emph{arXiv preprint arXiv:1604.07316}, 2016.

\bibitem[Vaswani et~al.(2017)Vaswani, Shazeer, Parmar, Uszkoreit, Jones, Gomez, Kaiser, and Polosukhin]{vaswani2017attention}
Ashish Vaswani, Noam Shazeer, Niki Parmar, Jakob Uszkoreit, Llion Jones, Aidan~N Gomez, {\L}ukasz Kaiser, and Illia Polosukhin.
\newblock Attention is all you need.
\newblock \emph{Advances in neural information processing systems}, 30, 2017.

\bibitem[Wo{\'z}niak et~al.(2022)Wo{\'z}niak, Wieczorek, and Si{\l}ka]{wozniak2022deep}
Marcin Wo{\'z}niak, Micha{\l} Wieczorek, and Jakub Si{\l}ka.
\newblock Deep neural network with transfer learning in remote object detection from drone.
\newblock In \emph{Proceedings of the 5th international ACM mobicom workshop on drone assisted wireless communications for 5G and beyond}, pages 121--126, 2022.

\bibitem[Ambalavanan et~al.(2020)]{ambalavanan2020cyber}
Vaishnavi Ambalavanan et~al.
\newblock Cyber threats detection and mitigation using machine learning.
\newblock In \emph{Handbook of research on machine and deep learning applications for cyber security}, pages 132--149. IGI Global, 2020.

\bibitem[Koch et~al.(2015)Koch, Zemel, Salakhutdinov, et~al.]{koch2015siamese}
Gregory Koch, Richard Zemel, Ruslan Salakhutdinov, et~al.
\newblock Siamese neural networks for one-shot image recognition.
\newblock In \emph{ICML deep learning workshop}, volume~2, page~0. Lille, 2015.

\bibitem[Vinyals et~al.(2016)Vinyals, Blundell, Lillicrap, Wierstra, et~al.]{vinyals2016matching}
Oriol Vinyals, Charles Blundell, Timothy Lillicrap, Daan Wierstra, et~al.
\newblock Matching networks for one shot learning.
\newblock \emph{Advances in neural information processing systems}, 29, 2016.

\bibitem[Snell et~al.(2017)Snell, Swersky, and Zemel]{snell2017prototypical}
Jake Snell, Kevin Swersky, and Richard Zemel.
\newblock Prototypical networks for few-shot learning.
\newblock \emph{Advances in neural information processing systems}, 30, 2017.

\bibitem[Finn et~al.(2017)Finn, Abbeel, and Levine]{finn2017model}
Chelsea Finn, Pieter Abbeel, and Sergey Levine.
\newblock Model-agnostic meta-learning for fast adaptation of deep networks.
\newblock In \emph{International conference on machine learning}, pages 1126--1135. PMLR, 2017.

\bibitem[Santoro et~al.(2016)Santoro, Bartunov, Botvinick, Wierstra, and Lillicrap]{santoro2016meta}
Adam Santoro, Sergey Bartunov, Matthew Botvinick, Daan Wierstra, and Timothy Lillicrap.
\newblock Meta-learning with memory-augmented neural networks.
\newblock In \emph{International conference on machine learning}, pages 1842--1850. PMLR, 2016.

\bibitem[Ravi and Larochelle(2016)]{ravi2016optimization}
Sachin Ravi and Hugo Larochelle.
\newblock Optimization as a model for few-shot learning.
\newblock In \emph{International conference on learning representations}, 2016.

\bibitem[Nichol et~al.(2018)Nichol, Achiam, and Schulman]{nichol2018first}
Alex Nichol, Joshua Achiam, and John Schulman.
\newblock On first-order meta-learning algorithms.
\newblock \emph{arXiv preprint arXiv:1803.02999}, 2018.

\bibitem[Raghu et~al.(2019)Raghu, Raghu, Bengio, and Vinyals]{raghu2019rapid}
Aniruddh Raghu, Maithra Raghu, Samy Bengio, and Oriol Vinyals.
\newblock Rapid learning or feature reuse? towards understanding the effectiveness of maml.
\newblock \emph{arXiv preprint arXiv:1909.09157}, 2019.

\bibitem[Bengio et~al.(1995)Bengio, Bengio, Cloutier, and Gecsei]{bengio1995optimization}
Samy Bengio, Yoshua Bengio, Jocelyn Cloutier, and Jan Gecsei.
\newblock On the optimization of a synaptic learning rule.
\newblock In \emph{Preprints Conf. Optimality in Artificial and Biological Neural Networks}, volume~2, 1995.

\bibitem[Hochreiter et~al.(2001)Hochreiter, Younger, and Conwell]{hochreiter2001learning}
Sepp Hochreiter, A~Steven Younger, and Peter~R Conwell.
\newblock Learning to learn using gradient descent.
\newblock In \emph{International Conference on Artificial Neural Networks}, pages 87--94. Springer, 2001.

\bibitem[Munkhdalai and Yu(2017)]{munkhdalai2017meta}
Tsendsuren Munkhdalai and Hong Yu.
\newblock Meta networks.
\newblock In \emph{International Conference on Machine Learning}, pages 2554--2563. PMLR, 2017.

\bibitem[Tiwari et~al.(2022)Tiwari, Gogoi, Verma, and Singh]{tiwari2022meta}
Sambhavi Tiwari, Manas Gogoi, Shekhar Verma, and Krishna~Pratap Singh.
\newblock Meta-learning with hopfield neural network.
\newblock In \emph{2022 IEEE 9th Uttar Pradesh Section International Conference on Electrical, Electronics and Computer Engineering (UPCON)}, pages 1--5. IEEE, 2022.

\bibitem[Sung et~al.(2018)Sung, Yang, Zhang, Xiang, Torr, and Hospedales]{sung2018learning}
Flood Sung, Yongxin Yang, Li~Zhang, Tao Xiang, Philip~HS Torr, and Timothy~M Hospedales.
\newblock Learning to compare: Relation network for few-shot learning.
\newblock In \emph{Proceedings of the IEEE conference on computer vision and pattern recognition}, pages 1199--1208, 2018.

\bibitem[Tang et~al.(2020)Tang, Li, Peng, and Tang]{tang2020blockmix}
Hao Tang, Zechao Li, Zhimao Peng, and Jinhui Tang.
\newblock Blockmix: meta regularization and self-calibrated inference for metric-based meta-learning.
\newblock In \emph{Proceedings of the 28th ACM international conference on multimedia}, pages 610--618, 2020.

\bibitem[Peng et~al.(2019)Peng, Li, Zhang, Li, Qi, and Tang]{peng2019few}
Zhimao Peng, Zechao Li, Junge Zhang, Yan Li, Guo-Jun Qi, and Jinhui Tang.
\newblock Few-shot image recognition with knowledge transfer.
\newblock In \emph{Proceedings of the IEEE/CVF international conference on computer vision}, pages 441--449, 2019.

\bibitem[Li et~al.(2023)Li, Tang, Peng, Qi, and Tang]{li2023knowledge}
Zechao Li, Hao Tang, Zhimao Peng, Guo-Jun Qi, and Jinhui Tang.
\newblock Knowledge-guided semantic transfer network for few-shot image recognition.
\newblock \emph{IEEE Transactions on Neural Networks and Learning Systems}, 2023.

\bibitem[Nichol and Schulman(2018)]{nichol2018reptile}
Alex Nichol and John Schulman.
\newblock Reptile: a scalable metalearning algorithm.
\newblock \emph{arXiv preprint arXiv:1803.02999}, 2\penalty0 (3):\penalty0 4, 2018.

\bibitem[Li et~al.(2017)Li, Zhou, Chen, and Li]{li2017meta}
Zhenguo Li, Fengwei Zhou, Fei Chen, and Hang Li.
\newblock Meta-sgd: Learning to learn quickly for few-shot learning.
\newblock \emph{arXiv preprint arXiv:1707.09835}, 2017.

\bibitem[Chen et~al.(2019)Chen, Liu, Kira, Wang, and Huang]{chen2019closer}
Wei-Yu Chen, Yen-Cheng Liu, Zsolt Kira, Yu-Chiang~Frank Wang, and Jia-Bin Huang.
\newblock A closer look at few-shot classification.
\newblock \emph{arXiv preprint arXiv:1904.04232}, 2019.

\bibitem[Tian et~al.(2020)Tian, Wang, Krishnan, Tenenbaum, and Isola]{tian2020rethinking}
Yonglong Tian, Yue Wang, Dilip Krishnan, Joshua~B Tenenbaum, and Phillip Isola.
\newblock Rethinking few-shot image classification: a good embedding is all you need?
\newblock In \emph{European Conference on Computer Vision}, pages 266--282. Springer, 2020.

\bibitem[Fan et~al.(2020)Fan, Lai, and Wang]{fan2020quasi}
Feng-Lei Fan, Rongjie Lai, and Ge~Wang.
\newblock Quasi-equivalence of width and depth of neural networks.
\newblock \emph{arXiv preprint arXiv:2002.02515}, 2020.

\bibitem[Nguyen et~al.(2020)Nguyen, Raghu, and Kornblith]{nguyen2020wide}
Thao Nguyen, Maithra Raghu, and Simon Kornblith.
\newblock Do wide and deep networks learn the same things? uncovering how neural network representations vary with width and depth.
\newblock \emph{arXiv preprint arXiv:2010.15327}, 2020.

\bibitem[Nguyen and Hein(2017)]{nguyen2017loss}
Quynh Nguyen and Matthias Hein.
\newblock The loss surface of deep and wide neural networks.
\newblock In \emph{International conference on machine learning}, pages 2603--2612. PMLR, 2017.

\bibitem[Lake et~al.(2011)Lake, Salakhutdinov, Gross, and Tenenbaum]{lake2011one}
Brenden Lake, Ruslan Salakhutdinov, Jason Gross, and Joshua Tenenbaum.
\newblock One shot learning of simple visual concepts.
\newblock In \emph{Proceedings of the annual meeting of the cognitive science society}, volume~33, 2011.

\bibitem[Oh et~al.(2020)Oh, Yoo, Kim, and Yun]{oh2020boil}
Jaehoon Oh, Hyungjun Yoo, ChangHwan Kim, and Se-Young Yun.
\newblock Boil: Towards representation change for few-shot learning.
\newblock \emph{arXiv preprint arXiv:2008.08882}, 2020.

\bibitem[Miranda et~al.(2021)Miranda, Wang, and Koyejo]{miranda2021does}
Brando Miranda, Yu-Xiong Wang, and Sanmi Koyejo.
\newblock Does maml only work via feature re-use? a data centric perspective.
\newblock \emph{arXiv preprint arXiv:2112.13137}, 2021.

\bibitem[Deleu et~al.(2019)Deleu, W\"urfl, Samiei, Cohen, and Bengio]{deleu2019torchmeta}
Tristan Deleu, Tobias W\"urfl, Mandana Samiei, Joseph~Paul Cohen, and Yoshua Bengio.
\newblock {Torchmeta: A Meta-Learning library for PyTorch}, 2019.
\newblock URL \url{https://arxiv.org/abs/1909.06576}.
\newblock Available at: https://github.com/tristandeleu/pytorch-meta.

\bibitem[Arnold et~al.(2021)Arnold, Iqbal, and Sha]{arnold2021maml}
S{\'e}bastien Arnold, Shariq Iqbal, and Fei Sha.
\newblock When maml can adapt fast and how to assist when it cannot.
\newblock In \emph{International Conference on Artificial Intelligence and Statistics}, pages 244--252. PMLR, 2021.

\end{thebibliography}





% \end{thebibliography}

\end{document}